\documentclass{article}

\usepackage{arxiv}

\usepackage[utf8]{inputenc} 
\usepackage[T1]{fontenc}    
\usepackage{hyperref}       
\usepackage{url}            
\usepackage{booktabs}       
\usepackage{amsfonts}       
\usepackage{nicefrac}       
\usepackage{microtype}      
\usepackage{graphicx}

\usepackage{epstopdf}
\usepackage{algorithmic}
\usepackage{algorithm}
\usepackage[footnotesize]{caption}
\usepackage{subcaption}
\usepackage{bbold}
\usepackage{multirow}
\usepackage{tikz}
\usepackage{pgfplots}
\usepackage{xcolor}
\usepackage{makecell}
\usepackage{adjustbox}
\usepackage{longtable}
\usepackage{amsmath}
\usepackage{cleveref}
\usepackage[margin=1cm]{caption}

\ifpdf
  \DeclareGraphicsExtensions{.eps,.pdf,.png,.jpg}
\else
  \DeclareGraphicsExtensions{.eps}
\fi

\newcommand{\norm}[1]{\left\lVert#1\right\rVert}
\DeclareMathOperator*{\minB}{min}

\newcommand{\argmax}[1]{\underset{#1}{\operatorname{argmax}}\;}
\newcommand{\argmin}[1]{\underset{#1}{\operatorname{argmin}}\;}

\DeclareMathOperator{\tr}{tr}

\title{Semi-supervised dictionary learning with graph regularization and active points}

\author{
 Khanh-Hung Tran \\
  Institut LIST, CEA \\
  Université Paris-Saclay\\
  Gif-sur-Yvette, 91191 \\
  \texttt{khanh-hung.tran@cea.fr} \\
   \And
Fred-Maurice Ngole-Mboula \\
  Institut LIST, CEA \\
  Université Paris-Saclay\\
  Gif-sur-Yvette, 91191 \\
  \texttt{fred-maurice.ngole-mboula@cea.fr} \\
  \And
Jean-Luc Starck \\
  Astrophysics Department, CEA\\
  Université Paris-Saclay\\
  Gif-sur-Yvette, 91191 \\
  \texttt{jean-luc.starck@cea.fr} \\
  \And
 Vincent Prost \\
  Institut LIST, CEA \\
  Université Paris-Saclay\\
  Gif-sur-Yvette, 91191 \\
  \texttt{vincent.prost@cea.fr} \\
}

\begin{document}
\maketitle
\begin{abstract}
Supervised Dictionary Learning has gained much interest in the recent decade and has shown significant performance improvements in image classification. However, in general, supervised learning needs a large number of labelled samples per class to achieve an acceptable result. In order to deal with databases which have just a few labelled samples per class, semi-supervised learning, which also exploits unlabelled samples in training phase is used. Indeed, unlabelled samples can help to regularize the learning model, yielding an improvement of classification accuracy. In this paper, we propose a new semi-supervised dictionary learning method based on two pillars: on one hand, we enforce manifold structure preservation from the original data into sparse code space using Locally Linear Embedding, which can be considered a regularization of sparse code; on the other hand, we train a semi-supervised classifier in sparse code space. We show that our approach provides an improvement over state-of-the-art semi-supervised dictionary learning methods. Several implementations of our work can be found in \textcolor{blue}{\url{https://github.com/ktran1/SSDL}}.
\end{abstract}


\section{Introduction}
\emph{Dictionary Learning} (DL) encompasses methods and algorithms that aim at deriving a set of primary features which enables one to concisely describe signals of a given type (see for example \cite{Aharon06, Mairal09_Online_DL, HLee07, barthelemy2012shift, ravishankar2011mr}). The benefit of such sparsity-driven dictionaries has been shown firstly in signal recovery applications such as denoising \cite{elad2006image,beckouche2013astronomical}, super-resolution \cite{Yang10SR, ngole14SR}, in-painting \cite{Zhang14IR} and colorization \cite{Mairal08colorization}. 

In the era of Machine Learning, DL use has been extended to classification tasks, yielding Supervised Dictionary Learning approaches (SDL). We can distinguish two categories of SDL methods: the ones which try to make sparse codes discriminative and those which try to make atoms discriminative. In the first category, a classifier which works in sparse code space is integrated into the classical DL problem \cite{Zhang10,Mairal09}. The second category relies on the relationship between atoms and class labels: a specific sub dictionary is learnt for each class and an unlabelled sample is classified according to the minimum reconstruction error sub dictionary \cite{Yang10, Wright09,Jiang13,Kong12}. A recent review of this field can be found in the work of Gangeh \textit{et al.} \cite{gangeh2015supervised}. 

Despite their merits, SDL methods have a critical shortcoming inherent to the supervised learning paradigm: training data is constrained by only labelled samples. Thus, if the number of labelled samples is not sufficient for constraining the dictionary, it might poorly capture discriminative features, yielding a low classification accuracy rate. Leveraging unlabelled data leads to a new framework of DL that we call Semi-Supervised Dictionary Learning (SSDL). On the one hand, SSDL allow for more data to be used in learning the dictionary, which in turn can better grasp common underlying signal structure of the data so that over-fitting can be avoided. On the other hand, unlabelled samples can also help to regularize in learning the classifier. More details will be provided in \cref{sec:Related work} which briefly introduces some related works in SSDL. 

We present our approach in \cref{sec:Proposed_Method} and deal with the optimization scheme in \cref{sec:Optimization}. The numerical experiments are presented and discussed in \cref{sec:Numerical_experiments}, followed by the conclusions and perspectives in \cref{sec:Conclusions}. Finally, in \hyperref[sec:Appendix]{Appendix}, we detail our method to optimize the Sparse Coding problem and provide some technical notes. 

\subsection*{Notations}
\label{subsec:Notations}
We adopt the following notation conventions :
\begin{itemize}
    \item lower case and bold letters are used for vectors;
    \item upper case and bold letters are used for matrices;
    \item $\mathbf{Z}[i,j]$ or $\mathbf{Z}_{i,j}$ is a component of $\mathbf{Z}$, where $i$ is the row number and $j$ is the column number;
    \item by default, vectors are represented as columns;
    \item Greek letters are used for hyper-parameters;
    \item non-bold letters are scalar.
    \item $\mathbb{0}, \mathbb{1}$ are matrices that contain only respectively 0 and 1. They have the same dimension as matrices or vectors associated in an operation.
    \item for a matrix $\mathbf{Z}$, we note $(\mathbf{Z}<\mathbb{1})$ the matrix $\mathbf{U}$ of the same size as $\mathbf{Z}$ satisfying :
$\mathbf{U}[i,j] =\begin{cases}
  1 \text{ if } \mathbf{Z}[i,j]<1 \\
  0 \text{ otherwise}
\end{cases}$

\end{itemize}
\medskip

The matrix $\mathbf{X} \in \mathbb{R}^{n\times N}$ denotes a set of $N$ samples $\mathbf{x}_i \in \mathbb{R}^n$, ($n$ : number of features). We assume that these data are made of two subsets: the labelled dataset $\mathbf{X}^{l}$ ($C$ classes) and the unlabelled dataset $\mathbf{X}^{u}$, which have respectively $N_{l}$ samples $\mathbf{x}^l_i$ $(i = 1,..,N_l)$ and $N_{u}$ samples $\mathbf{x}^u_j$ $(j = 1,..,N_u)$, hence $\mathbf{X} = [\mathbf{X}^{l},\mathbf{X}^{u}]$. The matrix $\mathbf{D} \in \mathbb{R}^{n\times p}$ denotes a dictionary which contains $p$ atoms $\mathbf{d}_{i} \in \mathbb{R}^{n}$ $\left(i=1,...,p \right)$ and $\mathcal{C} = \{ \mathbf{D}, \norm{\mathbf{d}_{i}}_2 \leq \alpha,  \forall i = 1,2,...,p \}$ is a subset of $\mathbb{R}^{n\times p}$ that contains all dictionaries whose atoms $l_2$ norms are less or equal to $\alpha$ (normally $\alpha = 1$).
\medskip

The matrix $\mathbf{A} \in \mathbb{R}^{p\times N}$  contains the sparse codes $\mathbf{a}_i$ for $\mathbf{x}_i$ in its columns. By analogy, $\mathbf{A}^{l}$ and $\mathbf{A}^{u}$ denote respectively the labelled and unlabelled sparse code matrices. Hence $\mathbf{A} = [\mathbf{A}^{l},\mathbf{A}^{u}]$. The labelled sparse code $\mathbf{a}_i^l$ is for the labelled sample $\mathbf{x}^{l}_i$ and the unlabelled sparse code $\mathbf{a}_j^u$ is for the unlabelled sample $\mathbf{x}^{u}_i$. %

\medskip
The vector $\mathbf{y}_i = [y_i^1,y_i^2,...,y_i^C]^\top \in \mathbb{R}^{C}$ indicates which class the
labelled sample $i$ belongs to, using the following convention: \\

$y_i^j =\begin{cases}
  1 \text{ if $i^{th}$ sample belongs to $j^{th}$ class} \\
  -1 \text{ otherwise}.
\end{cases}$

$\mathbf{Y} = [\mathbf{y}_1,\mathbf{y}_2,...,\mathbf{y}_{N_l}]$ is the label matrix for all labelled samples, $\mathbf{Y} \in \mathbb{R}^{C\times N_{l}}$.
\medskip

Finally, $\mathbf{W} \in \mathbb{R}^{C \times p}, \mathbf{W} = [\mathbf{w}_1,\mathbf{w}_2,...,\mathbf{w}_{C}]^\top$ is a linear classifier consisting of $C$ binary classifiers (with the strategy "one vs all") in the sparse code space and $\mathbf{b} = [b_1,b_2,...,b_C]^\top$ denotes the associated bias. 

\section{Related work}
\label{sec:Related work}

In this section, we outline some SSDL methods by presenting how a SDL version can be converted to SSDL and how we can improve further SSDL. Let's start with SDL in a general way with the following objective function :

\begin{equation}
\label{equa:objective_function_hidden_general_SDL}
\begin{split}
\minB_{\Theta}
&\left[\mathcal{R}(\mathbf{A}^l,\mathbf{D}) + \mathcal{D}(\mathbf{W},\mathbf{b},\mathbf{A}^l,\mathbf{D}) \right], \\
& \text{where $\Theta =  \{\mathbf{W},\mathbf{b},\mathbf{A}^l,\mathbf{D}\in \mathcal{C} \}$}.
\end{split}
\end{equation}

$\mathcal{R}$ denotes the reconstruction error with the sparsity constraint. The penalty $\mathcal{D}$ aims at making the dictionary $\mathbf{D}$ or the sparse codes $\mathbf{A}$ discriminative, possibly including an internal classifier ($\mathbf{W},\mathbf{b}$) learning loss. Here are two popular explicit objective functions following the form of \cref{equa:objective_function_hidden_general_SDL}. To classify unlabelled samples, the first one uses the learnt internal classifier. On the other hand the second one relies on reconstruction errors given by each specific sub dictionary.
\vspace{0.5cm}
\begin{minipage}{0.55\textwidth}
\begin{center}
\vspace{0.3cm}
\textbf{Objective function}
\end{center}
\end{minipage}%
\begin{minipage}{0.3\textwidth}
\begin{center}
\vspace{0.3cm}
\textbf{Prediction}
\end{center}
\end{minipage}%

\begin{minipage}{0.55\textwidth}
\begin{center}
\begin{equation*}
 \minB_{\mathbf{W},\mathbf{A}^l, \mathbf{D}\in \mathcal{C}} \underbrace{\norm{\mathbf{X}^l-\mathbf{D}\mathbf{A}^l}_{F}^{2} + \lambda\norm{\mathbf{A}^l}_q}_{\mathcal{R}} +\underbrace{\gamma\norm{\mathbf{Y}-\mathbf{W}\mathbf{A}^{l}}_{F}^{2}}_{\mathcal{D}}
\end{equation*}
\end{center}
\end{minipage}%
\begin{minipage}{0.35\textwidth}
\begin{equation}
\label{equation_explicite_of_SDL} 
\argmax{c} \mathbf{w}_c^\top \mathbf{a}^u  
\end{equation}
\end{minipage}%
\vspace{-0.5cm}
\begin{minipage}{0.55\textwidth}
\begin{equation*}
\end{equation*}
\end{minipage}%
\begin{minipage}{0.3\textwidth}
\begin{center}
\end{center}
\end{minipage}%


\begin{minipage}{0.55\textwidth}
\begin{center}
\begin{equation*}
\minB_{\mathbf{A}_{c,c}^l, \mathbf{D}\in \mathcal{C}} \underbrace{\sum\limits_{c=1}^{C}\Big(\norm{\mathbf{X}_{c}^l-\mathbf{D}_c\mathbf{A}_{c,c}^l}_{F}^{2} + \lambda \norm{\mathbf{A}_{c,c}^l}_q\Big)}_{\mathcal{R} \text{ and }\mathcal{D}}
\end{equation*}
\end{center}
\end{minipage}%
\begin{minipage}{0.35\textwidth}
\begin{center}
\begin{equation*}
\argmin{c} \norm{\mathbf{x}^u - \mathbf{D}_c \mathbf{a}^u }_2  
\end{equation*}
\end{center}
\end{minipage}%

\begin{minipage}{\dimexpr\textwidth-1cm}
\begin{center}
\vspace{0.3cm}
\end{center}
\end{minipage} 

\begin{minipage}{\dimexpr\textwidth-1cm}
\begin{center}
(where $\mathbf{D}_c, \mathbf{A}_{c,c}^{l}$ are respectively specific sub dictionary and sparse code to be learnt from the samples of $c^{th}$ class $\mathbf{X}_c^{l}$; $\mathbf{D} = [\mathbf{D}_1,..., \mathbf{D}_c, ..., \mathbf{D}_C]$ and $0\leq q \leq 1$)
\end{center}
\end{minipage} 
\vspace{0.3cm}

From the SDL objective function, we present three approaches : the first one is used to convert an SDL method into semi-supervised DL, i.e, to incorporate unlabelled samples in the learning, the second and third ones are used to reinforce semi-supervised learning.

The first straightforward way consists in modifying $\mathcal{R}$ by incorporating the reconstruction error and the sparse codes penalty for the unlabelled data: $\norm{\mathbf{X}^u-\mathbf{D}\mathbf{A}^u}_{F}^{2} + \lambda\norm{\mathbf{A}^u}_q$. This setting is found in most SSDL approaches (see for example \cite{Pham08,Zhang12, WangH13SSRD}).

The second one can go further by adding to the objective function a term $\mathcal{F}(\mathbf{A}^l,\mathbf{A}^u)$ to enforce the preservation of the manifold structure of the original data representation in the sparse code representation. Let consider these two manifolds are represented respectively by two graphs, one whose nodes are all available samples and other whose nodes are corresponding sparse codes. For preservation, a relation between samples is taken from first graph then it is applied to the second graph, meaning that the corresponding sparse codes need to respect this relation. The manifold structure preservation is applied in \cite{Zheng11, Yankelevsky17Mar} for sparse code regularization.

Finally, the third one can modify the functional $\mathcal{D}$ to make the supervised classification become semi-supervised, by using also unlabelled sparse codes as pseudo-labelled in $\mathcal{D}$. To do so, several works introduce a new matrix variable $\mathbf{P} \in \mathbb{R}^{C \times N_u}$, whose entry $\mathbf{P}_{kj}$ is positive and indicates the estimated probability that an unlabelled sample $j$ belongs to class $k$ (hence $\sum\limits_{k=1}^{C}\mathbf{P}_{kj} = 1$). Then, in some works, this matrix is used in the learning process to weight the internal classifier errors \cite{Wang15} or reconstruction errors \cite{Shrivastava12,Chen17} w.r.t. to each candidate class for unlabelled samples. $\mathbf{P}$ can be updated using different strategies.

Integrating the three aspects addressed above, we can formulate the following generic optimization problem for SSDL: 

\begin{equation}
\label{equa:objective_function_hidden_general}
\begin{split}
&\minB_{\Theta}\left[\mathcal{R}(\mathbf{A}^l,\mathbf{A}^u,\mathbf{D}) + \mathcal{D}(\mathbf{W},\mathbf{b},\mathbf{A}^l,\mathbf{A}^u,\mathbf{D},\mathbf{P}) + \mathcal{F}(\mathbf{A}^l,\mathbf{A}^u)\right], \\
&  \text{where $\Theta =  \{\mathbf{W},\mathbf{b},\mathbf{A}^l,\mathbf{A}^u,\mathbf{D}\in \mathcal{C},\mathbf{P} \}$}.
\end{split}
\end{equation}

The following table (table \ref{tab:SSDL comparaison}) represents explicit objective function of several semi-supervised dictionary learning methods, under form \cref{equa:objective_function_hidden_general}. Here are some new notations used in this table :

\begin{itemize}
    \item[-] $\mathbf{a}_{i,c}$ is sparse code on the sub-dictionary $\mathbf{D}_c$ for sample $\mathbf{x}_i$. Therefore $\mathbf{a}_i = [\mathbf{a}_{i,1}^\top;...;\mathbf{a}_{i,c}^\top;...;\mathbf{a}_{i,C}^\top]^\top$.
    \item[-] $\mathbf{A}_{k,c}^l$ is sparse code corresponds to the sub-dictionary $\mathbf{D}_c$ for the samples of $k^{th}$ class $\mathbf{X}^l_k$.
    \item [-] $\mathbf{A}_k^{l} = [\mathbf{A}_{k,1}^{l\top},..., \mathbf{A}_{k,c}^{l\top},..., \mathbf{A}_{k,C}^{l\top}]^\top $ and  $ \mathbf{A}^{l} = [\mathbf{A}^{l}_{1},...,\mathbf{A}^{l}_{k},...,\mathbf{A}^{l}_{C}]; \mathbf{X}^l = [\mathbf{X}^l_1,...,\mathbf{X}^l_k,...,\mathbf{X}^l_C]$.
    \item[-] $\mathrm{m}[\mathbf{Z}]$ is the mean vector (between columns) for matrix $\mathbf{Z}$ and $\mathrm{M}[\mathbf{Z}]$ is the matrix with the same size as $\mathbf{Z}$ by repeating column $\mathrm{m}[\mathbf{Z}]$. \item[-] $0 < q, q_1 < 1 $.
\end{itemize}

\begin{small}
\begin{longtable}[!ht]{|c|cccc|}
\hline
& & & &\\[\dimexpr-\normalbaselineskip+0.5em]
 Method &  & $\mathcal{R}$ & $\mathcal{D}$  & $\mathcal{F}$ \\
& & & &\\[\dimexpr-\normalbaselineskip+0.5em]
\hline
& & & &\\[\dimexpr-\normalbaselineskip+0.5em]
JDL \cite{Pham08} & \begin{tabular}{@{}c@{}} min  \\ $\mathbf{D}\in \mathcal{C} ,\mathbf{A},\mathbf{W} $\end{tabular}  & \begin{tabular}{@{}c@{}} $\norm{\mathbf{X}^l-\mathbf{D}\mathbf{A}^l}_F^2$  \\ $+\rho\norm{\mathbf{X}^u-\mathbf{D}\mathbf{A}^u}_F^2$ \\ \end{tabular}   & \begin{tabular}{@{}c@{}}$\gamma\norm{\mathbf{Y} - \mathbf{W}\mathbf{A}^l}_F^2$  \\ $+\mu\norm{\mathbf{W}}_F^2$ \\ \end{tabular} & \\
& s.t $\norm{\mathbf{a}_i}_0\leq \epsilon $ & & &  \\
& & & &\\[\dimexpr-\normalbaselineskip+0.5em]
\hline
& & & &\\[\dimexpr-\normalbaselineskip+0.5em]
 \begin{tabular}{@{}c@{}} OSSDL \cite{Zhang12}  \\  LC-RLSDLA \\ \cite{Matiz16,Irofti19}  \end{tabular} & \begin{tabular}{@{}c@{}} min  \\ $\mathbf{D}\in \mathcal{C} ,\mathbf{A},\mathbf{W},\mathbf{U}$ \end{tabular}  & \begin{tabular}{@{}c@{}} $\norm{\mathbf{X}^l-\mathbf{D}\mathbf{A}^l}_F^2$  \\ $+\rho\norm{\mathbf{X}^u-\mathbf{D}\mathbf{A}^u}_F^2$ \\ \end{tabular}   & \begin{tabular}{@{}c@{}}$\gamma\norm{\mathbf{Y} - \mathbf{W}\mathbf{A}^l}_F^2$  \\ $+\psi \norm{\mathbf{Q} - \mathbf{U}\mathbf{A}^l}_F^2$ \\ \end{tabular} & \\
& s.t $\norm{\mathbf{a}_i}_0\leq \epsilon $ & & & \\
& & & &\\[\dimexpr-\normalbaselineskip+0.5em]
\hline
& & & &\\[\dimexpr-\normalbaselineskip+0.5em]
SD2D \cite{Shrivastava12} & \begin{tabular}{@{}c@{}} min  \\ $\mathbf{D}\in \mathcal{C} ,\mathbf{A},\mathbf{P}$ \end{tabular}  & \begin{tabular}{@{}c@{}} $\norm{\mathbf{X}-\mathbf{D}\mathbf{A}}_F^2$  \\ $+ \lambda\norm{\mathbf{A}}_1$  \end{tabular}   & \begin{tabular}{@{}c@{}}$\sum\limits_{i = 1}^{N_l}\sum\limits_{c = 1}^{C}\norm{\mathbf{x}_i^l-\mathbf{D}_c\mathbf{a}_{i,c}^l}_2^2$  \\ $+\sum\limits_{j = 1}^{N_u}\sum\limits_{c = 1}^{C}\bigg(\norm{(\mathbf{x}^u_j-\mathbf{D}_c\mathbf{a}_{j,c}^u)\mathbf{P}_{cj}}_2^2$ \\$+\norm{\mathbf{D}_c\mathbf{a}_{j,c}^u(1-\mathbf{P}_{cj})}_2^2$ \bigg) \\
$+\sum\limits_{c = 1}^{C}\bigg(\norm{\mathbf{A}_c^l-\mathrm{M}[\mathbf{A}_{c}^{l}]}_F^2 $\\$-\norm{\mathrm{m}[\mathbf{A}^l_c]-\mathrm{m}[\mathbf{A}^l]}_2^2 \bigg) $\end{tabular} & \\
& & & &\\[\dimexpr-\normalbaselineskip+0.5em]
\hline
& & & &\\[\dimexpr-\normalbaselineskip+0.5em]
SSR-D * \cite{WangH13SSRD} & \begin{tabular}{@{}c@{}} min  \\ $\mathbf{D}\in \mathcal{C} ,\mathbf{A} $\end{tabular} & \begin{tabular}{@{}c@{}} $\norm{(\mathbf{X}-\mathbf{D}\mathbf{A})^\top}_{2,q_1}^{q_1}$ \\ $+ \lambda\sum\limits_{c=1}^C\norm{\mathbf{A}^l_c}_{2,q}^q$ \\ $+ \lambda\norm{\mathbf{A}^u}_{2,q}^q$    \end{tabular}   &  &  \\
& & & &\\[\dimexpr-\normalbaselineskip+0.5em]
\hline
& & & &\\[\dimexpr-\normalbaselineskip+0.5em]
SSP-DL * \cite{Wangdi16}  & \begin{tabular}{@{}c@{}} min  \\ $\mathbf{D}\in \mathcal{C} ,\mathbf{A} $\end{tabular} & \begin{tabular}{@{}c@{}} $\norm{\mathbf{X}-\mathbf{D}\mathbf{A}}_F^2$ \\ $+ \lambda_1\sum\limits_{c=1}^C\norm{\mathbf{A}^l_c}_{2,q}^q$ \\ $+ \lambda_2\norm{\mathbf{A}^u}_{q,q}^q$  \end{tabular}   &  & $\beta\norm{\mathbf{A} - \mathbf{A}\mathbf{V}}_F^2$ \\
& & & &\\[\dimexpr-\normalbaselineskip+0.5em]
\hline
& & & &\\[\dimexpr-\normalbaselineskip+0.5em]
USSDL \cite{Wang15}&  \begin{tabular}{@{}c@{}} min  \\ $\mathbf{D}\in \mathcal{C} ,\mathbf{A}$ \\ $\mathbf{W},\mathbf{b}, \mathbf{P}$ \end{tabular}  & \begin{tabular}{@{}c@{}} $\norm{\mathbf{X}-\mathbf{D}\mathbf{A}}_F^2$  \\ $+ \lambda\norm{\mathbf{A}}_1$  \end{tabular}  &\begin{tabular}{@{}c@{}} $\gamma\big(\sum\limits_{i=1}^{N_l}\sum\limits_{c=1}^{C}\norm{ \mathbf{w}_c^\top\mathbf{a}_i^l + b_c - y_i^c}_2^2$
  \\ $+ \sum\limits_{j=1}^{N_u}\sum\limits_{k=1}^{C}(\mathbf{P}_{kj})^r\sum\limits_{c=1}^{C}\norm{\mathbf{w}_c^\top\mathbf{a}_j^u + b_c -y_j^c(k) }_2^2 \big)$ \\ $+\mu\norm{\mathbf{W}}_F^2 $\end{tabular}
& \\
& & & &\\[\dimexpr-\normalbaselineskip+0.5em]
\hline
& & & &\\[\dimexpr-\normalbaselineskip+0.5em]
PSSDL \cite{Babagholami-Mohamadabadi13} & \begin{tabular}{@{}c@{}} min  \\ $\mathbf{D} ,\mathbf{A},\mathbf{W}$\\ $\sigma_i$,$z$,$\sigma_d$,$\sigma_w$ \end{tabular}  & \begin{tabular}{@{}c@{}} $\sum\limits_{i=1}^N\big(\frac{\norm{\mathbf{x}_i-\mathbf{D}\mathbf{a}_i}_2^2}{2\sigma_{i}^2}$ \\ $+ \log\sigma_i^{n+2} \big) + \frac{\norm{\mathbf{A}}_1}{z}$ \\ $+(N+1) \log z$  \\ $+\frac{\norm{\mathbf{D}}_F^2}{2\sigma_d^2}+p\log \sigma_d^{n+2}$\end{tabular}  & \begin{tabular}{@{}c@{}} $\sum\limits_{i}^{N_l}\big(-\mathbf{w}_{c,y^c_i=1}^\top \mathbf{a}_i^l$ \\$+ \log\big(\sum\limits_{c=1}^{C} exp(\mathbf{w}_{c}^\top\mathbf{a}_i^l\big)\big) $  \\ $+\frac{\norm{\mathbf{W}}_F^2}{2\sigma_w^2} + C\log\sigma_w^{p+2}$ \\ $ + (1-\beta)\big(\tr{(\mathbf{A}L^{LW}\mathbf{A}^\top)}$ \\$ - \tr{(\mathbf{A}L^{LB}\mathbf{A}^\top)}\big) $\end{tabular} & $\beta \tr{(\mathbf{A}L_A\mathbf{A}^\top)}$ \\
& & & &\\[\dimexpr-\normalbaselineskip+0.5em]
\hline
& & & &\\[\dimexpr-\normalbaselineskip+0.5em]
SSD-LP \cite{Chen17} & \begin{tabular}{@{}c@{}} min  \\ $\mathbf{D}\in \mathcal{C} ,\mathbf{A}, \mathbf{P}$ \end{tabular}  &
\begin{tabular}{@{}c@{}} $\sum\limits_{c=1}^{C}\big(\norm{\mathbf{X}_c^l - \mathbf{D}_c\mathbf{A}^l_{c,c}}_F^2 $  \\ $+\lambda\norm{\mathbf{A}^l_{c,c}}_1 \big)$ \end{tabular}  & \begin{tabular}{@{}c@{}} $-\gamma\sum\limits_{c=1}^{C}\norm{\mathbf{A}^l_{c,c} - \mathrm{M}[\mathbf{A}^l_{c,c}]}_F^2 $  \\ $+\sum\limits_{j}^{N_u} \sum\limits_{c=1}^{C}\big( \mathbf{P}_{cj}\norm{\mathbf{x}^u_j-\mathbf{D}_c\mathbf{a}^u_{j,c}}_2^2 + \lambda\norm{\mathbf{a}^u_{j,c}}_1\big)$\end{tabular} &\\
& & & & \\
\hline

\caption{\label{tab:SSDL comparaison} SSDL Objective functions. OSSDL and LC-RLSDLA are online learning methods, have the same objective function but they use the different methods for optimization : OSSDL uses ODL\cite{Mairal09_Online_DL} and LC-RLSDLA uses RLSDLA\cite{Skretting10}. In PSSDL, some hyper-parameters are learnt in optimization process. (*) These methods don't need label information while learning the dictionary, label information is used after that to classify firstly the atoms then unlabelled samples.} 

\end{longtable}
\end{small}

In our work, we construct our objective function based on the first objective function of \cref{equation_explicite_of_SDL}, because we believe that samples in the different classes have more or less similar textures and need to be reconstructed from the common dictionary. In addition, we want to train simultaneously the internal classifier and the dictionary to make sparse code more discriminative.

We propose the Semi-Supervised Dictionary Learning with Graph regularization and Active points method (SSDL-GA), which is extended version of USSDL with manifold structure preservation (see \cref{tab:SSDL comparaison}). This method also takes into account the manifold structure for sparse coding out-of-sample data points. We provide a detailed presentation in the next sections.

\section{Proposed Method}
\label{sec:Proposed_Method}


In our method, the reconstruction error and sparse coding term is simply $ \mathcal{R}(\mathbf{A},\mathbf{D}) = \norm{\mathbf{X} - \mathbf{D}\mathbf{A}}_F^2 + \lambda\norm{\mathbf{A}}_1$. It uses both labelled and unlabelled samples. 
The manifold structure preservation $\mathcal{F}$ is inspired by the Locally Linear Embedding (LLE)~\cite{Roweis00} method while the discrimination $\mathcal{D}$ relies on the internal semi-supervised classifier learning in the Adaptively Unified Classification approach~\cite{Wang15}. We detail in the following the corresponding functionals $\mathcal{F}$ and $\mathcal{D}$.     

\subsection{Manifold structure preservation}
\label{subsec:Manifold_structure_preservation} Assuming that the observed data is sampled from a smooth manifold and provided that the sampling is dense enough, one can assume that the data lie on locally linear manifold patches. Thus, LLE first computes the barycentric coordinates of the samples w.r.t. their nearest neighbors. These barycentric coordinates characterize the local geometry of the underlying manifold. Then, the LLE computes a low dimensional representation (an embedding) which is compatible with these local barycentric coordinates. 

We proceed with the same idea, considering the sparse codes as our embedding. Let knn($i$) denote a set containing indices of the k nearest neighbors samples (in Euclidean distance) of the sample $\mathbf{x}_i$, among the whole dataset, i.e. including both labelled and unlabeled samples. The barycentric coordinates of $\mathbf{x}_i$ w.r.t. its nearest neighbors are computed by solving the following optimization problem:
\begin{equation*}
\label{equa:LLE_sample}
\begin{split}
&\hat{\boldsymbol{\lambda}_{i}} = \minB_{\boldsymbol{\lambda}_{i} \in \mathbb{R}^{\text{k}} } \norm{\mathbf{x}_i - \sum\limits_{j\in \text{knn}(i)} \lambda_{ij} \mathbf{x}_{j}}_2^2,\\
&\text{subject to } \sum\limits_{j\in \text{knn}(i)} \lambda_{ij} = 1,\\
\end{split}
\end{equation*}
where $\boldsymbol{\lambda}_{i}$ is a vector of $k$ elements $\lambda_{ij}, j \in \text{knn}(i)$. Then we define $\mathcal{F}$ as:
\begin{equation*}
\label{equa:LLE_constraint_sparse_code}
\mathcal{F}(\mathbf{A}) =\beta \sum\limits_{i=1}^N \norm{\mathbf{a}_i - \sum\limits_{j\in \text{knn}(i)} \hat{\lambda}_{ij} \mathbf{a}_{j}}_2^2, 
\end{equation*}
where $\beta$ is a positive constant (hyper-parameter). 

Introducing the matrix $\mathbf{V} \in \mathbb{R}^{N \times N}$ as :
$\mathbf{V}[i,j] = \begin{cases} \hat{\lambda}_{ij} &\text{ if }  j \in \text{knn}(i) \\ 0 &\text{otherwise}\end{cases}$, $\mathcal{F}$ can be rewritten as $\mathcal{F}(\mathbf{A}) = \beta \norm{\mathbf{A} - \mathbf{A}\mathbf{V}}_F^2 = \beta  \tr(\mathbf{A}L_A\mathbf{A}^\top)$, where $L_A = \mathbf{I}_N-\mathbf{V}-\mathbf{V}^\top+\mathbf{V}^\top\mathbf{V}$. Under this form and $\mathbf{L}_A$ is a Laplacian matrix, the functional $\mathcal{F}$ can be interpreted as a graph Laplacian-based as in \cite{Zheng11, Yankelevsky16Dec}, but using an implicit metric for measuring distance between two samples.


\subsection{Adaptively Unified Classification with active points}
\label{subsec:Adaptatively Unified Classification}

For the functional $\mathcal{D}$, we construct this term in the same way as in approach USSDL \cite{Wang15}. Indeed, USSDL uses an internal semi-supervised classifier, which is developed from Adaptive Semi-Supervised Learning \cite{Wang14AdapSSL} and then combined with the Active points method. Following Adaptive Semi-Supervised Learning, $\mathcal{D}$ is split into two functionals $\mathcal{D}^l$ and $\mathcal{D}^u$ defined as: 
\begin{equation*}
\mathcal{D}^l(\mathbf{W},\mathbf{b},\mathbf{A}^l) = \gamma\sum\limits_{i=1}^{N_l}\sum\limits_{c=1}^{C}\norm{ \mathbf{w}_c^\top\mathbf{a}_i^l + b_c - y_i^c}_2^2 = \gamma\norm{\mathbf{W}\mathbf{A}^l + \mathbf{B}^l -\mathbf{Y}}_F^2,
\end{equation*}

and 
\begin{equation*}
\begin{split}
\mathcal{D}^u(\mathbf{W},\mathbf{b},\mathbf{A}^u,\mathbf{P}) & = \gamma \sum\limits_{j=1}^{N_u}\sum\limits_{k=1}^{C}(\mathbf{P}_{kj})^r\sum\limits_{c=1}^{C}\norm{\mathbf{w}_c^\top\mathbf{a}_j^u + b_c -y_j^c(k) }_2^2 \\
& = \gamma\sum\limits_{k=1}^{C} \norm{(\mathbf{P}_{k})^{r/2} \circ (\mathbf{W}\mathbf{A}^u + \mathbf{B}^u -\mathbf{Y}_k)}_F^2,\\
\end{split}
\end{equation*}
where :
\begin{itemize}
    \item[-] the matrices $\mathbf{B}^l$ and $\mathbf{B}^u$ consist respectively in $N_l$ and $N_u$ columns $\mathbf{b}$    
    \item[-] ($\circ$) is the Hadamard product
    \item[-] $\mathbf{P}_{k} \in \mathbb{R}^{C \times N_u}$ consists of $C$ rows equal to $\mathbf{P}[k,:]$  
    \item[-] $y_j^c(k) = 1$ if $k=c$, otherwise $y_j^c(k) = -1$. $\mathbf{y}_j(k) = [y_j^1(k),y_j^2(k),...,y_j^C(k)]^\top$ and $\mathbf{Y}_k = [\mathbf{y}_1(k),\mathbf{y}_2(k),...,\mathbf{y}_{N_u}(k)]$
    \item[-] $r \geq 1$ can be considered as activation hyper-parameter in function $x^r$, with $0 \leq x \leq 1 $.
\end{itemize}

As a reminder, $\mathbf{P}$ denotes a $C \times N_u$ matrix, whose entry $\mathbf{P}_{kj}$ is positive and indicates the estimated probability that an unlabelled sample $j$ belongs to class $k$, $N_u$ being the number of unlabelled samples. In $\mathcal{D}^u$, $\mathbf{P}$'s entries are used as weight parameters associated with classification error for unlabelled samples. 


The Active points method can be considered as the Hinge loss used in SVM. In classification problems, it can happen that some samples are classified in the true class but their sparse codes are far from the binary decision boundary (where $\mathbf{w}_c^\top\mathbf{a} + b_c = 0$) and the latter can be not optimal since it tries to fit with all sparse codes. To avoid this problem, only active points are used to rectify the decision boundary. Active points are just the sparse codes defined in the following way : $\mathbf{w}_c^\top\mathbf{a}_i+b_c < 1$ if $\mathbf{a}_i$ belongs to this class $c$ and $\mathbf{w}_c^\top\mathbf{a}_i+b_c > -1$ if not. More explanation and illustration can be found in USSDL \cite{Wang15}. Then, for labelled samples, matrix $\mathbf{Q}^l \in \mathbb{R}^{C \times N_l}$ indicates active points for a given class as follows : $\mathbf{Q}^l[c,i] = 1$ if $y_i^c( \mathbf{w}_c^\top\mathbf{a}_i^l + b_c) < 1$, otherwise $\mathbf{Q}^l[c,i] = 0$. In a similar fashion, we define the matrix $\mathbf{Q}^u_k \in \mathbb{R}^{C \times N_u}$, for unlabeled samples: $\mathbf{Q}^u_k[c,j] = 1$ if $y_j^c(k)(\mathbf{w}_c^\top\mathbf{a}_j^u + b_c) -1 < 0$, otherwise $\mathbf{Q}^u_k[c,j] = 0$,  $\forall k \in [1,..,C]$. Then $\mathcal{D}$ can be rewritten as follows:


\begin{equation*}
\begin{split}
&\mathcal{D} (\mathbf{W},\mathbf{b},\mathbf{A}^l,\mathbf{A}^u,\mathbf{P}) = \gamma\Big(\norm{\mathbf{Q}^l \circ (\mathbf{W}\mathbf{A}^l + \mathbf{B}^l -\mathbf{Y})}_F^2 \\
&+  \sum\limits_{k=1}^{C} \norm{\mathbf{Q}^u_k \circ (\mathbf{P}_{k})^{r/2} \circ (\mathbf{W}\mathbf{A}^u + \mathbf{B}^u -\mathbf{Y}_k)}_F^2 \Big) + \mu(\norm{\mathbf{W}}_F^2 + \norm{\mathbf{b}}_2^2),
\end{split}
\end{equation*}
where we have added a regularization on the linear classifier and the bias to avoid over-fitting since our model is trained with only few labelled samples (in \cref{sec:Numerical_experiments}).  

We try to represent $\mathcal{D}^l$ and $\mathcal{D}^u$ in the form of matrices to benefit from the fast operations between matrices in computation compared to iterating over $C,N_u,N_l$, which would be slow. By integrating the manifold structure preservation and the internal classifier learning terms, we end up with the following objective function:

\begin{equation}
\label{objective_function}
\begin{split}
\minB_{\mathbf{W},\mathbf{b},\mathbf{A}, \mathbf{P},\mathbf{D}\in \mathcal{C}} &\norm{\mathbf{X}-\mathbf{D}\mathbf{A} }_{F}^{2} + \lambda\norm{\mathbf{A}}_{1} + \beta\tr(\mathbf{A}L_{A}\mathbf{A}^\top) + \gamma\Big(\norm{\mathbf{Q}^l \circ (\mathbf{W}\mathbf{A}^l + \mathbf{B}^l -\mathbf{Y})}_F^2 \\
&+ \sum\limits_{k=1}^{C} \norm{\mathbf{Q}^u_k \circ (\mathbf{P}_{k})^{r/2} \circ (\mathbf{W}\mathbf{A}^u + \mathbf{B}^u -\mathbf{Y}_k)}_F^2\Big) + \mu(\norm{\mathbf{W}}_F^2 + \norm{\mathbf{b}}_2^2),
\end{split}
\end{equation}
where:
\begin{itemize}
    \item[-] $\mathbf{Q}^l = (\mathbf{Y} \circ (\mathbf{W}\mathbf{A}^l + \mathbf{B}^l) < \mathbb{1})$ and $\mathbf{Q}^u_k = (\mathbf{Y}_k \circ (\mathbf{W}\mathbf{A}^u + \mathbf{B}^u) < \mathbb{1})$, $\forall k \in [1,..,C]$
    \item[-] $\mathbf{P} \in \mathbb{R}^{C \times N_u}, \mathbf{P}_{kj} \in [0,1] \text{ and } \sum\limits_{k=1}^{C}\mathbf{P}_{kj} =1, \forall j$
    \item[-] $\mathbf{P}_{k} \in \mathbb{R}^{C \times N_u}$ is made by repeating $C$ times $\mathbf{P}[k,:]$ as rows   
\end{itemize}

We propose a minimization scheme in the following section.

\section{Optimization}
\label{sec:Optimization}

\subsection{Alternate update}

In the optimization process, the five following steps are repeated until convergence is reached: 

\textbf{Active elements update:}
\begin{equation*}
\begin{split}
&\mathbf{Q}^l = (\mathbf{Y} \circ (\mathbf{W}\mathbf{A}^l + \mathbf{B}^l) < \mathbb{1}) \\
&\mathbf{Q}^u_k = (\mathbf{Y}_k \circ (\mathbf{W}\mathbf{A}^u + \mathbf{B}^u) < \mathbb{1}),\forall k \in [1,..,C] \\
\end{split}
\end{equation*}

The complexity for this step is $\mathcal{O}(pCN_l + pC^2N_u)$.

\textbf{Probability update:}
\begin{equation*}
\begin{split}
&\minB_{\mathbf{P}\geq0}\sum\limits_{j=1}^{N_u}\sum\limits_{k=1}^{C}(\mathbf{P}_{kj})^r\sum\limits_{c=1}^{C}\mathbf{Q}^u_k[c,j]\norm{y_j^c(k)(\mathbf{w}_c^\top\mathbf{a}_j^u + b_c) -1 }_2^2, \\
& \text{ subject to } \sum\limits_{k=1}^{C} \mathbf{P}_{kj} = 1 , \forall j.
\end{split}
\end{equation*}

This is a convex optimization problem, given that $r\geq 1$. It can be solved efficiently using different methods depending on $r$ values. We use the same method as in \cite{Wang14AdapSSL}. The complexity for this step is $\mathcal{O}(pC^2N_u)$.

\textbf{Sparse coding :} 
\begin{equation}
\label{function_sp}
\begin{split}
\minB_{\mathbf{A}} \norm{\mathbf{X}-\mathbf{D}\mathbf{A}}_{F}^{2} + \lambda\norm{\mathbf{A}}_{1} & + \beta\tr(\mathbf{A}L_{A}\mathbf{A}^\top) + \gamma\Big(\norm{\mathbf{Q}^l \circ (\mathbf{W}\mathbf{A}^l + \mathbf{B}^l -\mathbf{Y})}_F^2 \\
& + \sum\limits_{k=1}^{C} \norm{\mathbf{Q}^u_k \circ (\mathbf{P}_{k})^{r/2} \circ (\mathbf{W}\mathbf{A}^u + \mathbf{B}^u -\mathbf{Y}_k)}_F^2\Big)
\end{split}
\end{equation}

The problem can be solved efficiently using FISTA with backtracking \cite{Beck09}. We give more details in \hyperref[sec:Appendix]{Appendix}. The complexity for this step is $\mathcal{O}((p^2N + pN^2 + npN+ pN_lC+pC^2N_u)s_s)$, where $s_s$ is number of iterations.  

\textbf{Dictionary update :} 
\begin{equation*}
\label{function_du}
\minB_{\mathbf{D} \in \mathcal{C}} \norm{\mathbf{X}-\mathbf{D}\mathbf{A} }_{F}^{2}
\end{equation*}

As in Sparse Coding, we use FISTA with backtracking is used to solve this problem. The complexity for this step is  $\mathcal{O}(p^2N + (p^2n + pNn)s_d)$, where $s_p$ is the number of iterations.


\textbf{Classifier update:} 
\begin{equation*}
\label{function_cu}
\begin{split}
\minB_{\mathbf{W},\mathbf{b}}
\gamma\Big(\norm{\mathbf{Q}^l \circ (\mathbf{W}\mathbf{A}^l + \mathbf{B}^l -\mathbf{Y})}_F^2
 + \sum\limits_{k=1}^{C} & \norm{\mathbf{Q}^u_k \circ (\mathbf{P}_{k})^{r/2} \circ (\mathbf{W}\mathbf{A}^u + \mathbf{B}^u -\mathbf{Y}_k)}_F^2\Big) \\
& + \mu(\norm{\mathbf{W}}_F^2 + \norm{\mathbf{b}}_2^2)
\end{split}
\end{equation*}

We use the same approach as in \cite{Wang15} to solve this quadratic optimization problem. The complexity of this step is $\mathcal{O}(p^2N_uC^2 +p^2N_lC)$.

\begin{algorithm}
\renewcommand{\thealgorithm}{1}
\caption{SSDL-GA}
\label{algorithm:SSDL-GA}
\begin{algorithmic}[1] 
\REQUIRE $\mathbf{X},\mathbf{Y}, \beta, \text{k}, \gamma, \lambda, \mu, p, r $.
\STATE \textbf{Initialize :} $L_A$ with k, $\mathbf{D}, \mathbf{A},  \mathbf{W}, \mathbf{b}, \mathbf{Y}_k $
\WHILE{not converged}
\STATE Update $\mathbf{Q}^l = (\mathbf{Y} \circ (\mathbf{W}\mathbf{A}^l + \mathbf{B}^l) < 1), \mathbf{Q}^u_k = (\mathbf{Y}_k \circ (\mathbf{W}\mathbf{A}^u + \mathbf{B}^u) < 1)$. 
\STATE Update the probability matrix $\mathbf{P}$ 
\STATE Update sparse code $\mathbf{A}$ (Sparse coding)
\STATE Update dictionary $\mathbf{D}$
\STATE Update classifier $\mathbf{W}, \mathbf{b}$
\ENDWHILE
\STATE \textbf{Output :}  $\mathbf{D}, \mathbf{A},  \mathbf{W}, \mathbf{b}, \mathbf{P}$
\end{algorithmic}
\end{algorithm}

The global optimization is summarized in \cref{algorithm:SSDL-GA}. In general, since $p < 10n$, the complexity for the global algorithm is $\mathcal{O}\big((n^2N+nN^2+nC^2N)s_ss_t + (n^3+n^2N)s_ds_t +n^2NC^2s_t\big)$, where $s_t$ is the number of iteration for global algorithm. As the computational cost is proportional to $N^2$ in sparse coding by adding manifold structure preservation, we develop in \hyperref[sec:Appendix]{Appendix} a sparse coding strategy with each batch of samples.    

Once we have the optimal $\mathbf{W}$ and $\mathbf{b}$, the unlabelled sample $\mathbf{a}_i^u$ is classified into the class $\hat{j}$ according to the following equation:
\begin{equation*}
\begin{split}
&\hat{j} = \argmax{j} \mathbf{w}_j^\top\mathbf{a}_i^u + b_j, \\
& \text{ where } \mathbf{w}_j^\top \text{ is } j^{th} \text{ row of } \mathbf{W}
\end{split}
\end{equation*}

\subsection{Initialization} 

The dictionary $\mathbf{D}$ is initialized as follows: if there are more atoms than labelled samples ($p>N_l$), all the labelled samples are used as initial atoms and the remaining initial atoms are selected randomly from the unlabelled samples. Otherwise, we select randomly a labelled sample for each class until we obtain $p$ samples. 
Then, each atom $\mathbf{d}_i$ is projected on the $l_2$ sphere of radius $\alpha (\mathbf{D} \in \mathcal{C})$. The sparse codes $\mathbf{A}$ are initialized by solving the following LASSO problem with the initial dictionary $\mathbf{D}$:
\begin{equation*}
\minB_{\mathbf{A}} \norm{\mathbf{X}-\mathbf{D}\mathbf{A} }_{F}^{2} + \lambda\norm{\mathbf{A}}_1
\end{equation*}

Finally, the linear classifier $\mathbf{W}$ and $\mathbf{b}$ is initialized using only the labelled sparse codes by solving the following problem:
\begin{equation*}
\begin{split}
&\minB_{\mathbf{W},\mathbf{b}} \gamma\norm{\mathbf{Y}-\mathbf{W}\mathbf{A}^l-\mathbf{B}^l}_{F}^{2} + \mu(\norm{\mathbf{W}}_F^2 + \norm{\mathbf{b}}_2^2) \\
= & \minB_{\mathbf{W}'} \gamma\norm{\mathbf{Y}-\mathbf{W}'\mathbf{A}^{l*}}_{F}^{2} + \mu\norm{\mathbf{W}'}_F^2, \\
\end{split}
\end{equation*}
where $\mathbf{W}' = \mathbf{[}\mathbf{W},\mathbf{b}\mathbf{]}\in \mathbb{R}^{C \times (p+1)}$ (we add $\mathbf{b}$ as a column after the last one of $\mathbf{W}$) and
 $\mathbf{A}^{l*} = \begin{bmatrix} 
\mathbf{A}^{l} &  \\
\mathbf{1} \\ 
\end{bmatrix}
$, where $\mathbf{1}$ is the vector one (we add $\mathbf{1}$ as a row after the last one of $\mathbf{A}^{l}$).

The solution $\hat{\mathbf{W}'}$ is given in closed-form is given by:
\begin{equation*}
\hat{\mathbf{W}'} = \mathbf{Y}(\mathbf{A}^{l*})^\top\left(\mathbf{A}^{l*}(\mathbf{A}^{l*})^\top +\frac{\mu}{\gamma} I \right)^{-1}
\end{equation*}

Note that hyper-parameters $\gamma$ and $\mu$ are the same in the initialization and in the optimization process. In our experiments, we fix $ \frac{\mu}{\gamma} = 2$ and it seems good for the accuracy rate.   

\subsection{Out-of-sample data points}
\label{subsec:New_coming_unlabelled_data_points}

We suppose that the dictionary $\mathbf{D}$ and the sparse codes $\mathbf{A} \in \mathbb{R}^{p \times N}$ have been calculated for $N$ training samples. 
If we have $q$ new unlabelled data points $\mathbf{X}^{new} =  [\mathbf{x}_{N+1},\mathbf{x}_{N+2},..., \mathbf{x}_{N+q}]$, we perform a simple sparse coding step for each new unlabelled data point $\mathbf{x}_{N+i}$, taking into account manifold structure preservation as follows:

\begin{equation}
\begin{split}
\label{equa:sparse_coding_new_unlabelled}
\minB_{\mathbf{a}_{N+i}} \norm{\mathbf{x}_{N+i}-\mathbf{D}\mathbf{a}_{N+i}}_{2}^{2} + \beta\norm{\mathbf{a}_{N+i} - \hat{\lambda}_{ij}\sum\limits_{j \in \text{knn}'(N+i)}\mathbf{a}_{j}}_2^2 +\lambda\norm{\mathbf{a}_{N+i}}_{1}.
\end{split}
\end{equation}

The set knn$'(N+i)$ contains the indices of the k nearest samples among the $N$ training samples for $\mathbf{x}_{N+i}$. We get the coefficients $\hat{\lambda}_{ij}$ by solving a problem of the form (\ref{equa:LLE_sample}), as previously mentioned. 


\section{Numerical experiments}
\label{sec:Numerical_experiments} We organize this section as follows : first, we show the advantage of the manifold structure preservation constraint on the USPS database (United States Postal Service). Then we assess the impact of the number of unlabelled samples involved in the training on both USPS and MNIST databases \cite{LeCun10}. Using the same datasets, we compare the performance of our approach with other SSDL methods, as well as Convolutional Neural Network (CNN) and Label Spreading (LP) classifiers. Finally, we evaluate our approach in the setting where very few labels are available using the two faces databases, Extended YaleB \cite{YaleB01} and AR \cite{Martinez98}. Note that data pre-processing is very important and will be detailed in each experiment. 

\subsection{Manifold structure preservation for sparse code regularization}
\label{subsec:Sparse code regularization}

The advantage of manifold structure preservation for regularizing sparse code in dictionary learning has been shown in several works \cite{Zheng11, Yankelevsky17Mar, Yankelevsky16Dec}. In this subsection, we evaluate the effect of the different Laplacian matrices $L_{A}$ (showed in \cref{tab:methods Laplacian different}) on the USPS handwritten digits dataset and their robustness to noise. 
This data is composed of 9298 images (16$\times$16), represented by 256-dimensional vectors. 
The training set only contains $N_l$ labelled samples which are extracted from 7291 training images and the testing set contains all 2007 testing images. In order to assess manifold structure preservation, we use the following objective function to obtain the dictionary and labelled sparse code : 
\begin{equation}
\label{equa:GraphSC}
\minB_{\mathbf{A}^{l}, \mathbf{D}\in \mathcal{C}} \norm{\mathbf{X}^{l}-\mathbf{D}\mathbf{A}^{l}}_{F}^{2} + \beta\tr{(\mathbf{A}^{l}L_{A}\mathbf{A}^{l\top})}  + \lambda\norm{\mathbf{A}^{l}}_{1}
\end{equation}

\begin{small}
\begin{longtable}{|l|l|}
\hline
 & \\[\dimexpr-\normalbaselineskip+0.5em]
 Method & Laplacian matrix $L_A$ and sparse coding for a testing sample $\mathbf{x}$ \\ [0.5ex] 
\hline
 & \\[\dimexpr-\normalbaselineskip+0.5em]
knn \cite{Zheng11} & \begin{tabular}{@{}l@{}} $W_{ij} = w(\mathbf{x}_i^l, \mathbf{x}_j^l) = \begin{cases} exp(\frac{-\norm{\mathbf{x}_i^l -\mathbf{x}^l_j}_2^2}{2\sigma^2}) , \text{if } \mathbf{x}_j^l\in \text{knn}(\mathbf{x}_i^l) \text{ or } \mathbf{x}_i^l\in \text{knn}(\mathbf{x}_j^l) \\ 0, \text{otherwise} \end{cases}$ \\ \rule{0pt}{4ex}  $L_A = D-W, \omega = \frac{N_l}{\tr{(L_A)}}, L_A \leftarrow \omega L_A$ \\[1ex] \hline \rule{0pt}{4ex} $ \minB_{\mathbf{a}} \norm{\mathbf{x}-\mathbf{D}\mathbf{a}}_{2}^{2}  + \lambda\norm{\mathbf{a}}_{1} + \beta \omega\sum_{j} \frac{1}{2}w(\mathbf{x},\mathbf{x}_j^l) \norm{\mathbf{a} - \mathbf{a}^{l}_{j}}^2_2 $ \\[1ex] \end{tabular} \\
\hline
 & \\[\dimexpr-\normalbaselineskip+0.5em]
 threshold \cite{Yankelevsky16Dec} & \begin{tabular}{@{}l@{}} $W_{ij} = w(\mathbf{x}_i^l, \mathbf{x}_j^l) = \begin{cases} exp(\frac{-\norm{\mathbf{x}_i^l -\mathbf{x}^l_j}_2^2}{2\sigma^2}) , \text{if } \norm{\mathbf{x}_i^l -\mathbf{x}^l_j}_2 < \kappa \\ 0, \text{otherwise} \end{cases}$ \\\rule{0pt}{4ex} $L_A = D-W, \omega = \frac{N_l}{\tr{(L_A)}}, L_A \leftarrow \omega L_A $ \\[1ex] \hline \rule{0pt}{4ex} $ \minB_{\mathbf{a}} \norm{\mathbf{x}-\mathbf{D}\mathbf{a}}_{2}^{2}  + \lambda\norm{\mathbf{a}}_{1} + \beta \omega\sum_{j} \frac{1}{2}w(\mathbf{x},\mathbf{x}_j^l) \norm{\mathbf{a} - \mathbf{a}^{l}_{j}}^2_2 $ \\[1ex] \end{tabular}  \\ 
\hline 
 & \\[\dimexpr-\normalbaselineskip+0.5em]
\begin{tabular}{@{}c@{}} Learnt Laplacian \\ \cite{Dong16} \end{tabular} & \begin{tabular}{@{}l@{}} $L_A = \minB_{L} \tr{(\mathbf{X}^lL\mathbf{X}^{l\top}) + \beta_{L}\norm{L}^2} $  \\ \rule{0pt}{4ex} s.t $ tr(L) = N_l, L_{ij} = L_{ji} < 0 (i \neq j$), $\Sigma_j L[i,j] = 0$ \\[1ex] \hline \rule{0pt}{4ex} $L_B = \minB_{L} \tr{([\mathbf{X}^l,\mathbf{x}]L[\mathbf{X}^l,\mathbf{x}]^\top) + \beta_{L}\norm{L}^2} $ \\\rule{0pt}{4ex} s.t $ tr(L) = N_l + 1, L_{ij} = L_{ji} < 0 (i \neq j$), $\Sigma_j L[i,j] = 0$  \\\rule{0pt}{4ex}
$\minB_{\mathbf{a}} \norm{\mathbf{x}-\mathbf{D}\mathbf{a}}_{2}^{2}  + \lambda\norm{\mathbf{a}}_{1} + \beta \big(2\tr{(\mathbf{a} L_B[N_l+1,1:N_l] \mathbf{A}^{l\top})} + \tr{(\mathbf{a}L_B[N_l+1,N_l+1]\mathbf{a}^\top)}\big) $ \\[1ex]\end{tabular} \\
\hline
 & \\[\dimexpr-\normalbaselineskip+0.5em]
LLE & \begin{tabular}{@{}l@{}} As in \cref{subsec:Manifold_structure_preservation}, then normalization : \\ \rule{0pt}{4ex} $L_A = D-W, \omega = \frac{N_l}{\tr{(L_A)}}, L_A \leftarrow \omega L_A $ \\[1ex] \hline \rule{0pt}{4ex} As in \cref{subsec:New_coming_unlabelled_data_points} (except $\beta \leftarrow \beta \omega$ ) \\[1ex] \end{tabular} \\
\hline
\caption{\label{tab:methods Laplacian different} Laplacian matrix $L_A$ given by different methods and the sparse coding that takes into account manifold structure preservation for testing sample. In all methods, each Laplacian matrix is normalized ($\tr{(L_A)} = N_l$) to have equal impact with the same hyper-parameter $\beta$.}
\end{longtable}
\end{small}

Since no internal classifier is learnt in the objective function \cref{equa:GraphSC}, the labelled sparse code is used to train an external linear SVM classifier. This classifier is tuned with different box constraint values $\{0.1, 1, 10 \}$, a "one against all" strategy and  five-fold cross validation. 
Then each testing sample is sparse coded with regularization as described in \cref{tab:methods Laplacian different}. Finally, the trained SVM predicts labels for testing sparse code. 
The experiment is performed as following. Firstly, we set $\beta = 0$ to find the best pair ($\lambda,p$), with $\lambda \in \{ 0.1,0.2,0.3,0.4,0.5,1 \} $ and $p \in \{ 32,64,128,256 \}$. Secondly, we fix the best pair ($\lambda,p$) = ($0.5,128$) to tune the remaining hyper-parameters (which depend on the method): $\beta, \beta_L, \sigma, \kappa, k$ (\cref{tab:value for hyper-parameters}). For $\kappa$, we introduce an additional hyper-parameter $\zeta$, which represents the percentile of distances such as $\norm{\mathbf{x}^l_i - \mathbf{x}^l_j}_2 < \kappa$. 
For $\sigma$, we note that the mean distance $\norm{\mathbf{x}^l_i - \mathbf{x}^l_j}_2$ in this dataset is about 10, which explains the selected range value  for $\sigma$. In all cases, we repeat three times with three random initializations for the dictionary and take the best score.          

\begin{table}[!ht]
\begin{center}
\begin{tabular}{|l|l|l|} 
\hline
\begin{tabular}{@{}c@{}} Hyper- \\ parameter \end{tabular}  & Values & Method \\
\hline
$\beta$ & $\{ 0.01,0.1,1,10,100\}$ & All \\
$k$ & $\{ 2,3,4,5,6,7,8,9 \}$ & knn, LLE \\
$\sigma$ & $\{0.1,1,10,15,30,1000\} $ & knn, threshold \\
$\zeta$ (for $\kappa$) & $\{0.03,0.05,0.1,0.15,0.3,0.5,0.7\} $ & threshold \\
$\beta_L$ & $\{ 10^{-3},5\times10^{-3},0.01,0.05,0.1,0.5,1,5,10,100\}$ & Learnt Laplacian \\
\hline
\end{tabular}
\end{center}
\caption{\label{tab:value for hyper-parameters}  Hyper-parameter value to select for sparse code regularization.}
\end{table}

Table \ref{tab:table USPS} shows the best error rates (with the best hyper-parameters) for each Laplacian matrix. Firstly, we see that using sparse code regularization ($\beta \neq 0$) gives better error rate than no using sparse code regularization ($\beta = 0$). Secondly, the Laplacian matrix given by LLE gives the best error rate. That explains our first choice to use this method in this paper. In addition, Laplacian matrix by LLE requires only one hyper-parameter, $k$, and therefore takes less time for tuning hyper-parameter task. Compared to Laplacian matrix given by method 'knn' and 'threshold', the one given by LLE is less dependent of the metric used to compute distance between samples (only need to determinate $k$ neighbor samples). The learnt Laplacian method does not require any metric and only depends on the penalty hyper-parameter $\beta_L$. Nevertheless, it is not trivial to interpret the geometric meaning of $L$. That is why, in this experiment, we need to find a new $L_B$ for each testing sample, which takes a lot of time when the number of samples is large. This motivated our decision not to perform the evaluation in cases $N_l$ = 1000 or 2000.

\begin{table}[!ht]
\begin{center}
\begin{tabular}{|l|c|c|c|c|c|c|} 
\hline
\multicolumn{1}{|c|}{Methods / $N_l$} & 100 & 500 & 1000 & 2000 \\ [0.5ex] 
 \hline
SC ($\beta = 0$) & 19.6 & 10.9 & 8.0 & 7.5  \\
 \hline
SC-knn & 16.6 & 9.1 &6.8 &6.0 \\ 
 \hline
SC-Threshold & 18.1 & 9.4 & 7.9 & 7.1 \\
 \hline
SC-Learnt Laplacian & 17.33 & 9.4 & & \\
 \hline
SC-LLE & 16.6 & 8.3 & 6.4 & 5.8 \\
\hline
\end{tabular}
\end{center}
\caption{\label{tab:table USPS} The error rate of classification on different types of Laplacian matrix used for sparse code regularization, with different number of labelled samples in training (same number of samples per class). SC means sparse coding.}
\end{table}

The last part of this subsection aims to evaluate the robustness of Laplacian matrix given different methods against additive noise in signal. We fix the number labelled training samples to 500 and add different noise amplitudes to both training and testing samples. The hyper-parameters are tuned as in previous experiment to get the best error rate for each method. \cref{tab:table USPS noise} shows these best error rates. From $\sigma = 0$ to $\sigma = 0.2$ (little noise), we see that the Laplacian matrix given by LLE is more sensitive to noise compared to the ones given by other methods. This observation is reasonable because LLE (without sparse coding) is known to be noise sensitive. Nevertheless, in general, SC-LLE gives good error rate among compared methods, this may be explained by the combination of sparse coding (which is robust to noise) and LLE. To deal with additive noise, we can use a simple trick that sets $\beta$ = 0 for some first iterations. Then learning $L_A$ with reconstructed samples $\mathbf{D}\mathbf{A}^l$, in which noise is reduced.

\begin{table}[!ht]
\begin{center}
\begin{tabular}{|l|c|c|c|c|c|c|c|c|} 
\hline
\multicolumn{1}{|c|}{Methods / $\sigma$} & 0 & 0.2 & 0.4 & 0.6 & 0.8 & 1 \\ [0.5ex] 
 \hline
SC ($\beta = 0$) & 10.9 & 10.9 & 11.8 & 12.7  & 15.7 & 17.6 \\
 \hline
SC-knn & 9.1 & 9.3 & 9.9 & 10.7 & 12.3 & 13.9\\ 
 \hline
SC-Threshold & 9.4 &9.5 &10.5 &11.9 &13.7 & 15.2\\
 \hline
SC-Learnt Laplacian & 9.4 & 9.5 & 9.8 & 10.5 & 12.0 & 14.6\\
 \hline
SC-LLE & 8.3 & 9.0 & 9.8 & 10.9 & 12.0 & 13.9\\
\hline
\end{tabular}
\end{center}
\caption{\label{tab:table USPS noise} The error rate of classification on different types of Laplacian matrix used for sparse code regularization. with different noise added amplitude. Here we fix the number of labelled samples in training $N_l = 500$ and use gaussian noise $\mathcal{N}(0,\sigma_N)$ where $\sigma_N = \sigma \mu(\mathbf{X}^{2})$.}
\end{table}

\subsection{Low number of labelled samples}
\label{subsec:Low_number_of_labelled_samples}

In this subsection, we evaluate our approach SSDL-GA in two tests on USPS and MNIST databases. MNIST contains images ($28 \times 28$) of 10 handwritten digits, 60000 images for the training set and 10000 images for the testing set. 

In the first test, for each database, we select for each class: 20 images as labelled samples, 100 images as testing samples and an increasing number of images 2, 5, 10, 20, 500, 100, 150 as unlabelled samples. 

Each image (as a vector) is normalized to have unit $l_2$ norm. Since we constrain $||\mathbf{d}_i||_2 \leq \alpha$ and we want to tune $\alpha$ for several values, an other way to do this is to fix $\alpha = 1$ and multiply normalized images by a scalar. Here we multiply normalized images by 5 which is equivalent to constrain $||\mathbf{d}_i||_2 \leq 0.2$. As mentioned before, sparse coding is performed for testing samples as in \cref{subsec:New_coming_unlabelled_data_points}.

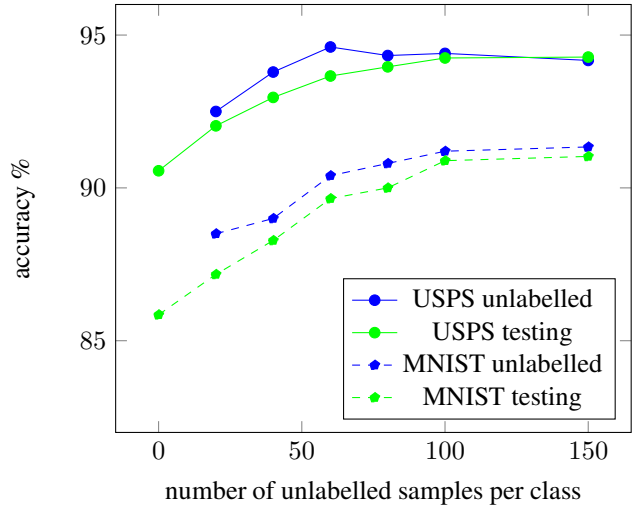
\begin{figure}
\centering
\begin{tikzpicture}
\begin{axis}
[ymax = 96, ymin = 82, xlabel = {number of unlabelled samples per class},
ylabel = {accuracy \%}, 
legend pos = south east]

\addplot [mark = *, color = blue] coordinates {
 (20,92.50)
(40,93.79) (60,94.61)(80,94.33)(100,94.4)(150,94.17)};
\addlegendentry{USPS unlabelled}

\addplot [mark = *, color = green] coordinates {
(0,90.56) (20,92.03)
(40,92.96) (60,93.66)(80,93.96)(100,94.25)(150,94.28)};
\addlegendentry{USPS testing}

\addplot [mark = pentagon*, color = blue, dashed] coordinates {
(20,88.5) (40,89)
(60,90.4) (80,90.8)(100,91.2)(150,91.34)};
\addlegendentry{MNIST unlabelled}

\addplot [mark = pentagon*, color = green, dashed] coordinates {
(0,85.85) (20,87.17) (40,88.28)
(60,89.65) (80,90)(100,90.89)(150,91.03)};
\addlegendentry{MNIST testing}

\end{axis}
\end{tikzpicture}
\caption{The accuracy rate for unlabelled samples and for testing samples in two databases USPS and MNIST with different number of unlabelled samples per class : 2,5,10,20,50,100,150.}
\label{fig:unlabelled_vs_testing}
\end{figure}

Five random samplings were conducted and the average scores for this test are shown in \cref{fig:unlabelled_vs_testing}. To tune hyper-parameters, we perform a grid search. We take $\mu =  2\gamma$ since $\mu$ is less sensitive compared to other hyper-parameters. As in previous subsection, we tune first for the pair $(\lambda ,p)$ while fixing $(\gamma = 0, \beta = 0)$. Then we fix the best found pair $(\lambda ,p)$ and tune for the remaining hyper-parameters. For the USPS database, we used the hyper parameters : $p = 200, \lambda = 0.3, \alpha = 1, \beta = 0.5, \gamma = 0.5, \mu = 1, \text{k} = 8, r= 1.7$ and for the MNIST database :  $p = 200, \lambda = 0.5, \alpha=1, \beta = 1, \gamma = 1, \mu = 2, \text{k} = 8, r = 2$. From this test we make two observations. First, the accuracy rate can be significantly improved by increasing the number of unlabelled samples in training and it converges to a stable value. This means that it is not necessary to use as many unlabelled samples as possible when the latter is numerous. Secondly, with a sufficient number of unlabelled samples in training, the accuracy rates for these unlabelled samples and for testing samples are the similar. From these two observations, we can notice that after a certain number of unlabelled samples, the model is well regularized. Therefore sparse code of testing samples, encoded as described in \cref{subsec:New_coming_unlabelled_data_points}, and encoded by retraining the model with labelled and existing unlabelled samples give the same performance. 

In the second test, we compare SSDL-GA with other SSDL approaches with the same pre-processing and hyper-parameters as in the first test. We set up the training set (labelled samples, unlabelled samples) and testing set as the same in \cite{Chen17}: 
\begin{itemize}
    \item For the MNIST database, we randomly select 200 images from each class, in which 20 images are used for labelled samples, 80 images are used for unlabelled samples and 100 images remain as testing samples. 
    \item For the USPS database, we randomly select 110 images from each class, in which 20 images are used for labelled samples, 40 images are used for unlabelled samples and 50 images remain as testing samples.
\end{itemize}

Five random samplings were conducted to calculate the mean and standard deviation on testing set. The \cref{tab:table MNIST USPS SSDL} shows the accuracy rate of various SSDL approaches : OSSDL \cite{Zhang12}, S2D2 \cite{Shrivastava12}, SSR-D \cite{WangH13SSRD}, SSP-DL \cite{Wangdi16},USSDL \cite{Wang15}, PSSDL \cite{Babagholami-Mohamadabadi13}, SSD-LP \cite{Chen17}, also CNN (supervised) and LP (semi-supervised). For CNN, just labelled samples are used for training and the shown results are the best average accuracy rate after trying with three different CNN models. Here are configurations used for MNIST and USPS respectively :

\textit{Conv}[$32\times 3\times 3$] $\rightarrow$ \textit{ReLU} $\rightarrow$ \textit{BNorm} $\rightarrow$ \textit{Conv}[$32\times 3\times 3$] $\rightarrow$ \textit{ReLU} $\rightarrow$ \textit{Pool}[$2\times 2$] $\rightarrow$ \textit{BNorm} $\rightarrow$ \textit{Conv}[$64\times 3\times 3$] $\rightarrow$ \textit{ReLU} $\rightarrow$ \textit{BNorm} $\rightarrow$ \textit{Conv}[$64\times 3\times 3$] $\rightarrow$ \textit{ReLU} $\rightarrow$ \textit{Pool}[$2\times 2$] $\rightarrow$ \textit{BNorm} $\rightarrow$ \textit{FC}[512] $\rightarrow$ \textit{ReLU} $\rightarrow$ \textit{BNorm} $\rightarrow$ \textit{Dropout}[0.25] $\rightarrow$ \textit{FC}[10] $\rightarrow$ \textit{softmax}.

\textit{Conv}[$16\times 3\times 3$] $\rightarrow$ \textit{ReLU} $\rightarrow$ \textit{BNorm} $\rightarrow$ \textit{Conv}[$32\times 3\times 3$] $\rightarrow$ \textit{tanh} $\rightarrow$ \textit{Pool}[$2\times 2$] \textit{BNorm} $\rightarrow$ \textit{FC}[128]  $\rightarrow$  \textit{tanh} $\rightarrow$ \textit{BNorm} $\rightarrow$ \textit{Dropout}[0.25] $\rightarrow$ \textit{FC}[10] $\rightarrow$ \textit{softmax}.

In both databases, first, we see the effect of manifold structure preservation with LLE by comparing SSDL-GA and USSDL (which is exactly SSDL-GA with $\beta = 0$). 
Second, the SSDL-GA outperforms other SSDL methods, as well as the CNN and LP. We notice that on MNIST, SSDL-GA is the only one in the SSDL family that can perform better than the CNN. This can be trivially explained by the fact that one can use unlabelled samples in training and the other can not. Therefore, we evaluate this test with a semi-supervised neural network model that can also use unlabelled samples in training, for example Ladder net \cite{Rasmus15}. Here is configuration used in Ladder net : \textit{FC}[1000] $\rightarrow$ \textit{ReLU} $\rightarrow$ \textit{FC}[500] $\rightarrow$ \textit{ReLU} $\rightarrow$ \textit{FC}[250] $\rightarrow$ \textit{softmax} and $\sigma_{noise} = 0.3$.
We see that our model gets slightly better accuracy rate compared to Ladder net.

\begin{table}[!ht]
\begin{center}
\begin{tabular}{|c|c|c|l|} 
\Xhline{3\arrayrulewidth}
Method / Data & USPS & MNIST & Sparse coding  \\ [0.5ex] 
\Xhline{3\arrayrulewidth}

LP & $90.3 \pm 1.3$ & $85.12 \pm 0.6$  & \\

\hline

OSSDL* & $80.8 \pm 2.8$ & $73.2 \pm 1.8$ & individual, $l_0$ \\
 
SD2D* & $86.6 \pm 1.6$ & $77.6 \pm 0.8$ & group, $l_1$ \\
 
SSR-D* & $87.2 \pm 0.5$ & $83.8 \pm 1.2$ & individual, $l_{2,p}$  \\
 
SSP-DL* & $87.8 \pm 1.1$ & $85.8 \pm 1.2$& group, $l_{2,p}$ and $l_{p,p}$   \\
 
USSDL &  $ 91.56 \pm 1.15 $ &  $84.8 \pm 1.7 $ & individual, $l_1$ \\
 
PSSDL $^\diamond$ & $86.9 \pm 1.0$ & $87.4 \pm 1.2$ &group, $l_1$ \\
 
SSD-LP* & $90.3 \pm 1.3$ & $87.8 \pm 1.6$&group, $l_1$  \\
 
SSDL-GA &  $\mathbf{93.6} \pm \mathbf{1.0}$ &  $\mathbf{90} \pm \mathbf{0.8}$ &group, $l_1$ \\

\hline
CNN & $89.28 \pm 1.4$ & $88.4 \pm 1.1$  &\\
\hline
Ladder  & $ 92.68 \pm 1.0$ & $89.84 \pm 0.8$ & \\
\Xhline{3\arrayrulewidth}

\end{tabular}
\end{center}
\caption{\label{tab:table MNIST USPS SSDL} Accuracy rate and nature of sparse coding for various semi-supervised methods, with handwritten digits databases USPS and MNIST. ($^\diamond$) In PSSDL, for each class, we use 25 images as labelled samples instead of 20, the rest of the training data as unlabelled samples and all testing data for testing samples; its corresponding accuracy rate is extracted from the original paper. (*) Accuracy rate are extracted from \cite{Chen17}.}
\end{table}

\subsection*{Complexity comparison}

As the complexity for several algorithms (in \cref{tab:SSDL comparaison}) is not communicated and we do not dispose the implementation for these algorithms, we compare then the complexity by each step in the optimization process. First of all, all objective functions of SSDL algorithms are iteratively optimized by following steps : sparse coding, dictionary update, classifier update and probability update. The first two steps (solved by iterative method) are the most essential and influence mostly to the complexity. The remaining steps can be solved trivially by first order optimality. Second, dictionary update is slightly different between the compared algorithms (which depends on the shared dictionary approach or the specific class sub-dictionaries approach). However, in general, we minimize a quadratic optimization problem $\norm{\mathbf{X}-\mathbf{D}\mathbf{A}}_2^2 $ for $\mathbf{D}$ with the constraint $\mathbf{D}\in \mathcal{C}$, therefore, the complexity of this step can be regarded as equivalent among the compared algorithms. By two observations, sparse coding step is the essential factor for the complexity comparison. 

The complexity of sparse coding then depends essentially by two factors : the type of sparse coding and the norm used in sparse coding. First, there are two types of sparse coding for a sample : individual sparse coding which do not depends on others sparse codes and group sparse coding, in the contrary, depends on others sparse codes. The first one is parallelizable but the second one is not because there are interactions between sparse code. In \cref{tab:SSDL comparaison}, sparse coding in the approaches that use manifold structure preservation $\mathcal{F}$ or Fisher Discriminant Analysis are group sparse coding. On the contrary, sparse coding in remaining approaches are individual sparse coding. Second, there are several norms used in sparse coding  $l_0;l_1;l_{2,q};l_{q,q}; (0<q<1)$. Norm $l_0$ has lower complexity (by MP-based) than the remaining norms (by gradient-based algorithm), which have equivalent complexity. In conclusion, for complexity comparison, we see first the type of sparse coding (group sparse coding has higher complexity than individual sparse coding), then we see the norm used in sparse coding. These two factors are showed for each SSDL method in \cref{tab:table MNIST USPS SSDL}, and we can see clearly the trade-off between complexity and accuracy.

\subsection{Face databases}

In this subsection, we evaluate our approach with Extended YaleB cropped and AR cropped dataset for which the size of each image is respectively $192 \times 168$ and $165 \times 120$ pixels.

The YaleB database contains 2432 frontal-face images of 38 individuals (64 images for each individual), captured under various illumination conditions and expressions. We first resize images to $54\times48$ before applying a Principal Component Analysis (PCA) to obtain 300 dimensional feature vectors (same process as \cite{Wang15,Chen17}). 
Then each vector coordinate is normalized to have zero mean and unit variance. Finally, each vector is normalized to have $l_2$ unit norm and then multiplied by 2. We randomly select $N=20$ images for each person to create a training set and use the remaining images for the testing set. 
In the training set, for each person, we use $N_l = \{2,5,10\}$ images as labelled samples and the remaining images as unlabelled samples. Five independent evaluations were conducted to compute the mean and standard deviation. 
The results are shown in \cref{table YaleB} with various SSDL approaches. In the case $N_l = 2$ (very few labelled samples), our approach improves significantly the accuracy rate compared to other SSDL methods but it is less accurate in the cases $N_l = \{5,10\}$. It is possible that the hyper-parameters are still not optimal for these cases.  
The hyper-parameters values in three cases are $N_l = \{2,5,10\}$ : $p = 380, \lambda = 0.005 , \beta = 2, \gamma = 0.5, \mu = 1, \text{k} = 4, r= 1.5$.

The AR Face database consists of over 4000 images but we evaluate our approach with its subset that consists of 2600 images (26 images per person for 50 male subjects and 50 female subjects). These 26 images are taken from different facial expressions, illumination conditions, and occlusions (sun-glasses and scarves). First, this database is projected onto a 540-dimensional feature vector by a randomly generated matrix as \cite{Jiang13, Yankelevsky17Mar}. The preprocessing is the same as described above as in the YaleB analysis.
For each person, 15 images are randomly selected for labelled set and 5 images are randomly selected for unlabelled set, which gives 20 images as training set and 6 images for testing set. The results shown in \cref{table AR} are extracted from \cite{Yankelevsky17Mar} for two SDL methods : LC-KSVD and SupGraphDL, which differ from our approach 
by using a training set that contains only labelled images (20 samples). 
Although our approach uses fewer labelled samples in the training set, it gives a better accuracy rate. 
The hyper-parameters used are $p = 300, \lambda = 0.0015 , \beta = 0.3, \gamma = 0.08, \mu = 0.016, \text{k} = 8, r= 1.5$.


\begin{table}[!ht]
\begin{center}
\begin{tabular}{cccc} 
\Xhline{3\arrayrulewidth}
Method / $N_l$ & 2 & 5 & 10 \\ [0.5ex] 
 \hline
S2D2 & $53.4 \pm 2.1$  & $76.1 \pm 1.3$ &  $83.2 \pm 1.9$\\

JDL & $55.2 \pm 1.8$  & $77.4 \pm 2.8$ &  $85.3 \pm 1.6$ \\

USSDL & $60.5 \pm 2.1$  & $86.5 \pm 2.1$ &  $93.6 \pm 0.8$ \\

SSD-LP & $67.0 \pm 2.9$  & $\mathbf{89.8} \pm \mathbf{0.9}$ &  $\mathbf{95.2} \pm \mathbf{0.2}$ \\

SSDL-GA & $\mathbf{73.62} \pm \mathbf{3.1}$  & $ 86.6 \pm 1.6 $ & $90.7 \pm 0.4$  \\
\hline

\Xhline{3\arrayrulewidth}
\end{tabular}
\vspace{0.2cm}
\captionof{table}{Accuracy rate for YaleB database with various SS-DL approaches and different number of labelled samples in training.}
\label{table YaleB}
\end{center}
\end{table}

\begin{table}[!ht]
\begin{center}
\begin{tabular}{|c|c|} 
\Xhline{3\arrayrulewidth}
Method  & Accuracy rate  \\ [0.5ex] 
\hline
LC-KSVD1 & $84.17 $ \\

LC-KSVD2 & $85 $ \\

SupGraphDL & $84.93 $ \\

SupGraphDL-L & $85.33 $ \\
\hline
SSDL-GA & $\mathbf{92.09} \pm \mathbf{1.16} $ \\
\Xhline{3\arrayrulewidth}
\end{tabular}
\vspace{0.2cm}
\captionof{table}{Accuracy rate for the AR database with SDL methods.}.
\label{table AR}
\end{center}
\end{table}

\section{Conclusions}
\label{sec:Conclusions}

We have presented a SSDL method by integrating manifold structure preservation and an internal semi-supervised classifier to the classical DL problem. This helps to exploit more information from unlabelled samples to reinforce the model. In addition, new unlabelled samples are also sparse coded by taking into account manifold structure preservation. Experimental results on several benchmark databases have shown the advantage of our approach, especially in the case of and low number of unlabelled samples in training, it performs about $2\%$ better than the state-of-art for digit recognition compared to other SSDL approaches and gets slightly better accuracy compared to semi-supervised neural network. We also propose a batch and epoch version in the appendix to accelerate the optimization process. However, in general, dictionary learning methods for classification objectives, due to limits of computation, require a dimensionality reduction to fewer than about $10^3$ dimensions, but applying dimensionality reduction can make an important loss of discriminatory information. Possible future work includes using a patch method to tackle this problem and dealing also with geometric invariance to have a more efficient model.

\newpage
\bibliographystyle{unsrt}
\bibliography{references}

\newpage
\section*{Appendix}
\label{sec:Appendix}

\subsection*{Sensibility for hyper-parameters}

\begin{itemize}
\item We tested also some large value for $k \in \{ 16,24,32 \}$ but $k < 10$ seems pertinent for our experiments. This is reasonable since if $k$ is too large, it happens that a data point has always some neighboring points which are center points of original manifold (graph), then the corresponding sparse codes tend to centralize and have regular distribution (loss of original manifold information).  

\item $r > 1$ in activation function $x^r$ for class probability of unlabelled samples. This function $x^r, 0 \leq x \leq 1$ helps to deactivate weak probabilities, but if $r$ is too large, $x^r$ deactivates also strong probabilities. Then we use these activated probability as pseudo labelled to train the classifier. The accuracy rate for testing samples is sensitive to this hyper-parameter, we need to tune it carefully (in our test we take values from 1 to 2 with step 0.1).   

\item $\lambda $ and $p$ are the two essential hyper-parameters for dictionary learning. These hyper-parameters are selected to ensure that a sample can be reconstructed by a linear combination of several atoms. The number of atom $p$ must be bigger the number of hidden dimensions of data (redundant).  

\end{itemize}

\subsection*{Sparse coding}
\label{subsec:Sparse_coding}

To apply FISTA, we separate (\ref{function_sp}) into two functions $f_{1}^{sp}$ and $ f_{2}^{sp}$:

\begin{alignat*}{2}
& f_{1}^{sp}(\mathbf{A}) = \lambda\norm{\mathbf{A}}_{1} \\
& f_{2}^{sp}(\mathbf{A}) = \norm{\mathbf{X}-\mathbf{D}\mathbf{A} }_{F}^{2} & + \beta\tr{(\mathbf{A}L_{A}\mathbf{A}^\top)} + \gamma\norm{\mathbf{Q}^l \circ (\mathbf{W}\mathbf{A}^l + \mathbf{B}^l -\mathbf{Y})}_F^2 \\
& & + \gamma\sum\limits_{k=1}^{C} \norm{\mathbf{Q}^u_k \circ (\mathbf{P}_{k})^{r/2} \circ (\mathbf{W}\mathbf{A}^u + \mathbf{B}^u -\mathbf{Y}_k)}_F^2
\end{alignat*}

The gradient of $f_{2}^{sp}$ is given by :

\begin{alignat*}{3}
\nabla f_{2}^{sp}(\mathbf{A}) &= -2\mathbf{D}^\top\left(\mathbf{X} - \mathbf{D}\mathbf{A}\right) + 2\beta\left(\mathbf{A}L_{A} \right) + 2\gamma \bigg[ \mathbf{W}^\top \Big( (\mathbf{Q}^l)^2\circ (\mathbf{W}\mathbf{A}^{l}  - \mathbf{B}^l - \mathbf{Y})\Big), \\
&\sum\limits_{k=1}^{C} \mathbf{W}^\top \Big((\mathbf{Q}^u_k)^2 \circ (\mathbf{P}_k)^r \circ (\mathbf{W}\mathbf{A}^{u}  - \mathbf{B}^u - \mathbf{Y}_k) \Big) \bigg],
\end{alignat*}

where $(\mathbf{Q}^l)^2 = \mathbf{Q}^l \circ \mathbf{Q}^l$, $(\mathbf{Q}^u_k)^2 = \mathbf{Q}^u_k \circ \mathbf{Q}^u_k$ 

\begin{algorithm}
\renewcommand{\thealgorithm}{2}
\caption{Sparse coding (FISTA with backtracking)}
\label{alg:sparse coding}
\begin{algorithmic}[1] 
\REQUIRE $\mathbf{X}, \mathbf{A}_0,\mathbf{D}, \mathbf{W},\mathbf{b}, \mathbf{Y}, \mathbf{Y}_k, L_{A}, \beta, \gamma, \lambda,\mathbf{Q}^l,\mathbf{Q}^u_k$.
\STATE \textbf{Initialize :} $ \mathbf{Z}_0 \leftarrow \mathbf{A}_0, t_0 \leftarrow 1, \tau > 0, \eta >1 $.
\FOR{$n=0,1,...$}
\WHILE{ True}
\STATE $\mathbf{H} \leftarrow \mathbf{A}_n - \tau^{-1}\nabla f_{2}^{sp}(\mathbf{A}_n)$
\STATE $\mathbf{Z}_{n+1} \leftarrow \text{sign}(\mathbf{H}) \circ \text{max}(|\mathbf{H}| - \tau^{-1}\lambda,0) $
\IF{$f_{2}^{sp}(\mathbf{Z}_{n+1}) \leq f_{2}^{sp}(\mathbf{A}_{n}) + \big\langle \mathbf{Z}_{n+1}-\mathbf{A}_{n}, \nabla f_{2}^{sp}(\mathbf{A}_n) \big\rangle + \frac{\tau}{2}\norm{\mathbf{Z}_{n+1}-\mathbf{A}_{n}}^2$}
\STATE \textbf{break}
\ENDIF
\STATE $\tau \leftarrow \eta \tau$
\ENDWHILE
\STATE $t_{n+1} \leftarrow \frac{1+\sqrt{4 t_n^2 + 1}}{2}$
\STATE $\upsilon \leftarrow 1 + \frac{t_n - 1}{t_{n+1}}$
\STATE $ \mathbf{A}_{n+1} \leftarrow \mathbf{Z}_n + \upsilon(\mathbf{Z}_{n+1}-\mathbf{Z}_n) $
\ENDFOR
\end{algorithmic}
\end{algorithm}

FISTA with backtracking does not require a Lipschitz coefficient for $\nabla f_{2}^{sp}$ but anyway we try to calculate it to know relatively which factors that step descent depends on. We consider that $\norm{.}$ is the Euclidean norm. By using these following inequalities that apply to two matrices $E$ and $F$, $\norm{EF} \leq \norm{E}\norm{F}$, $\norm{E + F} \leq \norm{E} + \norm{F}$, $\norm{E\circ F} \leq \norm{F}$ (if $ \mathbb{0} \leq E \leq \mathbb{1}$ ) and note that $\mathbb{0} \leq \mathbf{Q}^l,\mathbf{Q}^u_k,\mathbf{P}_k^r \leq \mathbb{1} $, we prove that:
\begin{alignat*}{3}
 \lVert \nabla f_{2}^{sp}(\mathbf{A}_1) & - \nabla f_{2}^{sp}(\mathbf{A}_2) \rVert \\
& \leq 2 \lVert \mathbf{D}^\top\mathbf{D}(\mathbf{A}_1 - \mathbf{A}_2) \rVert  + 2\beta \lVert(\mathbf{A}_{1} - \mathbf{A}_{2})L_{A} \rVert  \\
& + 2\gamma \lVert \mathbf{W}^\top \rVert \left\lVert \left[\mathbf{W}(\mathbf{A}_{1}^l - \mathbf{A}_{2}^l),\sum\limits_{k=1}^C \mathbf{W}(\mathbf{A}_{1}^u - \mathbf{A}_{2}^u) \right] \right\rVert \\
& \leq 2 \lVert \mathbf{D}^\top\mathbf{D} \rVert \lVert \mathbf{A}_1 - \mathbf{A}_2 \rVert  + 2\beta \lVert\mathbf{A}_{1} - \mathbf{A}_{2} \rVert \lVert L_{A} \rVert  \\
& + 2\gamma \lVert \mathbf{W}^\top \rVert \lVert \mathbf{W} \rVert \sum\limits_{k=1}^C \left\lVert \left[(\mathbf{A}_{1}^l - \mathbf{A}_{2}^l),(\mathbf{A}_{1}^u - \mathbf{A}_{2}^u) \right] \right\rVert \\
& \leq 2 \left(\norm{\mathbf{D}^\top\mathbf{D}} + \beta \norm{L_{A}} + \gamma C\norm{\mathbf{W}}^2 \right) \norm{\mathbf{A}_1 - \mathbf{A}_2}
\end{alignat*}

The step descent is proportional to $\tau^{-1} = \frac{1}{2} \left(\norm{\mathbf{D}^\top\mathbf{D}} + \beta \norm{L_{A}} + \gamma C\norm{\mathbf{W}}^2 \right)^{-1}$ and the greater $\tau$ is, the more time we need to optimize.

\subsection*{Technical notes}
\label{subsec:Technical notes}

We employ FISTA which is an iterative method to solve the Sparse Coding and Dictionary Update problems so it is necessary to control the step descent although the latter is automatically found with backtracking search trick. First, in Sparse Coding, the step descent depends on $\norm{\mathbf{D}^\top\mathbf{D}},\norm{L_A},\beta,\gamma,C,\norm{\mathbf{W}}^2$ so our algorithm might converge very slowly if one of these values is too large. If we want to promote manifold structure preservation for sparse code, it is better to increase k (the number of neighbors samples) instead of $\beta$. But if k is too large or training data point are not regularly sampled, it happens that some data points (in the center of manifold) are always in the neighborhood of others, therefore some columns of $\mathbf{V}$ are not sparse and $\norm{L_A} \approx \norm{\mathbf{V}^\top\mathbf{V}}$ becomes large. Second, in Dictionary Update, the step descent depends on $\norm{\mathbf{A}\mathbf{A}^\top}$, therefore $\mathbf{A}$ needs to be sparse, meaning $\lambda$ need to be large enough. In our work, if $\lambda$ is too small, we use \cite{HLee07} instead of FISTA with backtracking. By experience via tests executed in our paper, SSDL-GU converges in fewer than 20 iterations. We fix a maximum of 50 iterations for Sparse Coding and Dictionary Update.

In case of large number of classes $C$, the number of training samples (labelled and unlabelled) can be large, we can employ the batch and epoch strategy to minimize the Sparse Coding problem. For example, we repeat $n$ epochs for $m$ batches divided from $\mathbf{A}$. Note that only manifold structure preservation expression needs to be slightly modified since it has the interaction between sparse codes via the matrix $L_A$. We define $\mathbb{U}$ as the set that contains all indices of the training samples, thus $\mathbb{U} =\{1,2,..,N\} $. In an epoch, for each batch $i$, we select randomly $n_i$ training samples ($n_i \approx N/m$) whose indices are stored in the set $M_i$. $M_i^C$ is the complement of $M_i$ in $\mathbb{U}$. We rewrite :    

\begin{alignat*}{3}
&\tr{(\mathbf{A}L_{A}\mathbf{A}^\top)} \\
&= \tr{\bigg(\big[\mathbf{A}[:,M_i],\mathbf{A}[:,M_i^C]\big]
\begin{bmatrix} 
L_{A}[M_i,M_i] & L_{A}[M_i,M_i^C] \\
L_{A}[M_i,M_i^C]^\top & L_{A}[M_i^C,M_i^C]
\end{bmatrix}\big[\mathbf{A}[:,M_i],\mathbf{A}[:,M_i^C]\big]^\top\bigg)} \\
&= \tr{\bigg(\big[\mathbf{A}_i,\mathbf{A}_i^C]\big]
\begin{bmatrix} 
L_{A[ii]} & L_{A[iC]} \\
L_{A[iC]}^\top & L_{A[CC]}
\end{bmatrix}\big[\mathbf{A}_i,\mathbf{A}_i^C]\big]^\top\bigg)} \\
& = \tr{(\mathbf{A}_i L_{A[ii]} \mathbf{A}_i^\top)} + 2 \tr{(\mathbf{A}_i L_{A[iC]} \mathbf{A}_i^{C\top})} + \tr{(\mathbf{A}_i^{C} L_{A[CC]} \mathbf{A}_i^{C\top})},
\end{alignat*}
where :
\begin{itemize}
    \item[-] $L_{A[ii]} =  L_{A}[M_i,M_i]$ is a $\mathbb{R}^{n_i \times n_i}$ matrix formed by extracting from $L_{A}$, rows $M_i$ and columns $M_i$.
    \item[-] $L_{A[CC]}$ is a $\mathbb{R}^{(N-n_i) \times (N-n_i)}$ matrix formed by extracting from $L_{A}$ rows $M_i^C$ and columns $M_i^C$.
    \item[-] $L_{A[iC]}$ is a $\mathbb{R}^{(n_i) \times (N-n_i)}$ matrix formed by extracting from $L_{A}$ rows $M_i$ and columns $M_i^C$.
\end{itemize}

Then we optimize for each $\mathbf{A}_i$ while fixing other batches ($\mathbf{A}_i^C$) as in the FISTA method for Sparse Coding mentioned before but this time we need to adjust the gradient of $2 \tr{(\mathbf{A}_i L_{A[iC]} \mathbf{A}_i^{C\top})}$ in the new $\nabla f^{sp}_2$. At the end of a batch, we update the sparse code before performing next batch. We suggest also evaluating the objective function after each step, as $\mathbf{Q}^l$, $\mathbf{Q}^u_k$ are fixed in an iteration, the objective function must decrease. Since $\mathbf{Q}^l$, $\mathbf{Q}^u_k$ are updated at the beginning of the next iteration, the objective function can increase slightly, but in general, we must see something decreases and converges.

\end{document}